\documentclass[ijoc, nonblindrev, final]{informs3} 

\OneAndAHalfSpacedXII 

\usepackage{natbib}
 \bibpunct[, ]{(}{)}{,}{a}{}{,}%
 \def\bibfont{\small}%
\TheoremsNumberedThrough     

\EquationsNumberedThrough    

\MANUSCRIPTNO{JOC-2023-03-SI-0055.R1}

\usepackage{booktabs}
\usepackage{multirow}
\usepackage{longtable}
\usepackage{caption}
\usepackage{listings}
\usepackage{pythonhighlight}

\usepackage[xcolor={divpdf,grey}, authormarkup=none]{changes}

\begin{document}

\renewcommand*{\lstlistlistingname}{Listings}
\renewcommand*{\lstlistingname}{Listing}

\newcommand{\relocate}{\texttt{RELOCATE}}
\newcommand{\swap}{\texttt{SWAP}}
\newcommand{\twoopt}{\texttt{2-OPT}}
\newcommand{\sstar}{\texttt{\textsuperscript{*}}}
\newcommand{\swapstar}{\swap\sstar}
\newcommand{\relocatestar}{\relocate\sstar}
\newcommand{\twooptstar}{\twoopt\sstar}


\RUNAUTHOR{Wouda, Lan, and Kool}

\RUNTITLE{PyVRP: a high-performance VRP solver package}

\TITLE{PyVRP: a high-performance VRP solver package}

\ARTICLEAUTHORS{%
\AUTHOR{Niels A. Wouda}
\AFF{Department of Operations, University of Groningen, Groningen, The Netherlands, \EMAIL{n.a.wouda@rug.nl}}
\AUTHOR{Leon Lan}
\AFF{Department of Mathematics, Vrije Universiteit Amsterdam, Amsterdam, The Netherlands, \EMAIL{l.lan@vu.nl}}
\AUTHOR{Wouter Kool}
\AFF{ORTEC, Zoetermeer, The Netherlands, \EMAIL{wouter.kool@ortec.com}}
} 

\ABSTRACT{%
We introduce PyVRP, a Python package that implements hybrid genetic search in a state-of-the-art vehicle routing problem (VRP) solver.
The package is designed for the VRP with time windows (VRPTW), but can be easily extended to support other VRP variants.
PyVRP combines the flexibility of Python with the performance of C++, by implementing (only) performance critical parts of the algorithm in C++, while being fully customisable at the Python level.
PyVRP is a polished implementation of the algorithm that ranked 1st in the 2021 DIMACS VRPTW challenge and, after improvements, ranked 1st on the static variant of the EURO meets NeurIPS 2022 vehicle routing competition.
The code follows good software engineering practices, and is well-documented and unit tested.
PyVRP is freely available under the liberal MIT license.
Through numerical experiments we show that PyVRP achieves state-of-the-art results on the VRPTW and capacitated VRP.
We hope that PyVRP enables researchers \added{and practitioners} to easily and quickly build on a state-of-the-art VRP solver.
}%


\HISTORY{}

\maketitle

%


\section{Introduction}
\label{sec:introduction}

This paper describes PyVRP, a Python package that \replaced{provides}{contains} a high-performance implementation of the hybrid genetic search (HGS) algorithm for vehicle routing problems (VRPs)~\citep{vidal2013hybrid}.
PyVRP currently supports two well-known VRP variants: the capacitated VRP (CVRP) and the VRP with time windows (VRPTW) \citep{toth2014vehicle}. 
The implementation builds on the open-source HGS-CVRP implementation of \citet{vidal2022hybrid}, but has added support for time windows and has been completely redesigned to be easy to use as a highly customisable Python package, while maintaining speed and state-of-the-art performance.

While HGS-CVRP is implemented completely in C++, PyVRP adopts a different philosophy: only performance-critical parts of the algorithm are implemented in C++, whereas all other parts are implemented in Python.
Besides easily using the algorithm, this gives the user additional flexibility to customise the algorithm using Python.
The provided \replaced{HGS}{Hybrid Genetic Search}\ implementation is one example of a large family of VRP algorithms that can be built using the building blocks that PyVRP provides.
As the performance-critical parts remain in C++, the added flexibility and ease-of-use that Python offers barely impact the algorithm performance.

PyVRP comes with pre-compiled binaries for Windows, Mac OS and Linux, and can be easily installed \replaced{from}{through PyPI,} the Python package index using \texttt{pip install pyvrp}.
This allows users to directly solve VRP instances, or implement variants of the HGS algorithm using Python, inspired by the examples in PyVRP's documentation. 
Users can customise various aspects of the algorithm using Python, including population management, crossover strategies, granular neighbourhoods and operator selection in the local search.
Additionally, for advanced use cases such as supporting additional VRP variants, users can build and install PyVRP directly from the source code.
We actively welcome community contributions to develop support for additional VRP variants within PyVRP, \added{and provide some guidelines for this in our online documentation}.

The goal of PyVRP, which is made available under the liberal MIT license, is to provide an easy-to-use, extensible, and well-documented VRP solver that generates state-of-the-art results on a variety of VRP variants. 
This can be used by practitioners to solve practical problems, and by researchers as a starting point or strong baseline when aiming to improve the state of the art. 
The name `PyVRP' has deliberately been chosen to not mention a specific algorithm or VRP variant, \replaced{providing flexibility with respect to the underlying heuristic algorithms and supported VRP variants}{While currently HGS is the state of the art, this may change and nothing prevents us from implementing another algorithm in the future, the implementation of which can reuse components already implemented in PyVRP}. 

Through the Python ecosystem, we enable a wide audience to easily use the software. 
We especially hope that PyVRP will help machine learning (ML) researchers interested in vehicle routing to easily build on the state-of-the-art, and move beyond LKH-3 \citep{helsgaun2017extension} as the most commonly used baseline \citep{accorsi2022guidelines}. 
Using ML for vehicle routing problems is a promising and active research area, but so far it has not been able to advance the state-of-the-art; this may change when ML researchers can build on a flexible and high-quality implementation of a state-of-the-art VRP solver.

\deleted{Compared to the HGS-CVRP code,}PyVRP is a complete Python library for solving multiple VRP variants, accompanied by unit tests, online documentation and examples on its use cases.
The library has been tested in practice: earlier versions of the software ranked 1st in the VRPTW track of the 12th DIMACS implementation challenge \citep{kool2022hybrid} and, after improvements, ranked 1st on the static VRPTW variant of the EURO meets NeurIPS 2022 vehicle routing competition \citep{van_doorn_solving_2022}.
Compared to the versions used to win these challenges, the PyVRP implementation has been simplified \replaced{further, and significantly rewritten to improve its overall design and testability.
In particular,}{and} complex components with limited contribution to the overall performance have been removed to strike a balance between simplicity and performance.
The result is simpler and more robust than the individual challenge solvers, which were quite complex and optimised for the specific challenge problem sets.
In Section~\ref{sec:experiments}, we show that PyVRP still yields excellent performance, using numerical experiments on public benchmarks following the conventions used in the DIMACS implementation challenge.

The rest of this paper is structured as follows.
In Section~\ref{sec:problem_description} we briefly introduce the CVRP and VRPTW, and the benchmarking conventions PyVRP supports.
In Section~\ref{sec:related_projects}, we discuss several other open-source VRP solvers, and highlight the novelty of PyVRP.
Then, in Sections~\ref{sec:technical_implementation} and~\ref{sec:package}, we discuss the technical implementation and PyVRP package, respectively.
In Section~\ref{sec:package} we \replaced{present two examples}{also present a short example} to demonstrate how the package can be used.
Section~\ref{sec:experiments} presents PyVRP's performance on \added{established} benchmark instances.
Finally, Section~\ref{sec:conclusion} concludes the paper.

\section{Problem description}
\label{sec:problem_description}

PyVRP currently supports two VRP variants: CVRP and VRPTW.
The CVRP aims to construct multiple routes, each starting and ending at the same depot, to serve a set of customers while minimising total distance.
The total demand of customers in a single vehicle is limited by the vehicle capacity. 
The VRPTW generalises the CVRP by adding the constraint that each customer must be visited within a certain time window.
\added{
For both CVRP and VRPTW, PyVRP supports minimising distances, not the number of vehicles used.
A simple procedure to support fleet minimisation can be developed on top of PyVRP, by first finding a feasible solution with some number of vehicles, and then removing one vehicle at a time until no feasible solution is found.
}
PyVRP has been designed to handle instances of these problems with up to several thousand customers.

\subsection{CVRP}
Formally, the capacitated VRP consists of customers $i = 1, ..., n$ with demands $q_i \ge 0$, which must be served from a common depot denoted as $0$.
The goal is to visit all customers using a fixed fleet of vehicles, each of which starts at, and returns to, the depot, while minimising the total distance travelled, where the distance from customer (or depot) $i$ to $j$ is denoted as $d_{ij} \ge 0$.
The total demand in each vehicle should not exceed the vehicle capacity $Q > 0$.

\subsection{VRPTW}
For the VRPTW, each customer additionally has a service time $s_i \ge 0$, an earliest arrival time $e_i \ge 0$ and latest arrival time $l_i \ge 0~(e_i \le l_i)$ in between which service should \emph{start}.
A vehicle can wait at customer $i$ when arriving too early, but cannot arrive after $l_i$.
The time to travel from customer (or depot) $i$ to $j$ is given by $t_{ij} \ge 0$.
\added{While in academic benchmarks the duration $t_{ij}$ is typically set equal to the distance $d_{ij}$, PyVRP supports separate distance and duration matrices, as is commonly encountered in practice.}

\subsection{Conventions}
There are different conventions on the definitions of the constraints and objectives for CVRP and VRPTW, especially relating to rounding of (Euclidean) distances and other data \added{in existing benchmark instances} (see e.g.\ \citet{uchoa2017new}). 
PyVRP supports \emph{integer} distances \added{and durations}, which should be provided \emph{explicitly} by the user.
\added{Additionally, PyVRP can be compiled to use double precision data as well, but that is not enabled by default for performance reasons: we found during initial experimentation that working with double precision data is somewhat slower than using integers.}
\added{For working with benchmark instances,} we rely on the \textsc{VRPLIB} package \added{\citep{lan_vrplib_2023}} to compute distances \added{and durations}, and we provide helper functions to scale and then round or truncate them before converting to integers.
This way, we support various conventions, including the one-decimal precision used in the DIMACS VRPTW challenge, and the CVRPLIB benchmark repository at \url{http://vrp.galgos.inf.puc-rio.br/}.

\section{Related projects}
\label{sec:related_projects}

As many algorithms developed in the literature have not been open-sourced, the primary goal of PyVRP is to open-source a state-of-the-art VRP solver that is easy to use and customise.
\replaced{We are aware of the following, related projects:}{There are a number of related projects that we would like to mention:}
\begin{itemize}
    \item HGS-CVRP \citep{vidal2022hybrid} is a simple, \replaced{specialised}{yet state-of-the-art} solver for the CVRP implemented in C++ with \replaced{state-of-the-art}{excellent} performance. 
    While a Python interface, PyHygese~\added{\citep{kwon_pyhygese_2022}}, is available on the Python package index, it does not come with pre-compiled binaries, and thus requires that users have a compiler toolchain already installed.
    Beyond setting the parameters, all types of customisation require changes of the C++ source code.
    \added{HGS-CVRP is open to contributions from the community and has a permissive MIT license.}

    \item LKH-3 \citep{helsgaun2017extension} is a \replaced{heuristic}{solver} that supports a wide variety of VRP \deleted{problem} variants. 
    It solves these by transforming them into a symmetric \replaced{travelling salesman problem}{TSP} problem and applying the Lin-Kernighan-Helsgaun local search heuristic.
    \replaced{While it provides good solutions for many problem variants, the solver is hard to customise as it requires modifying its C source code.}
    {While it has delivered best known solutions for multiple problem variants, it is currently not state-of-the-art for CVRP and VRPTW.}
    \added{Furthermore, it is only available under an academic and non-commercial license, and it is unclear whether community contributions are welcomed.}

    \item VROOM \added{\citep{vroom_v1.13}}, the Vehicle Routing Open-source Optimisation Machine, is an open-source solver that aims to provide good solutions to real-life VRPs.
    \added{In particular, it integrates well with open-source routing software to solve real-life VRPs within limited computation time.}
    It implements many constructive heuristics and a local search algorithm \added{in C++} and can handle different types of \replaced{VRPs}{problems}.
    \replaced{However, it is unable to compete with state-of-the-art algorithms and lacks documentation to customise its underlying solver.}{Although designed to provide good solutions quickly, it is unable to achieve state-of-the-art performance on CVRP and VRPTW even with longer runtimes.}

    \item OR-Tools \citep{ortools} is a general modelling and optimisation toolkit for solving operations research \deleted{(OR)} problems, \added{developed and} maintained by Google. 
    It is written in C++ but can be used from Python, Java or C\#. 
    \added{OR-tools is extensively documented and can be installed directly from the Python packaging index.
    Internally, it uses a constraint programming approach specialised to solve a large variety of routing problems. 
    While this approach allows it to model and solve many problem variants, its performance is far from the state of the art.}
    \deleted{While it is flexible and supports many problems, its performance is far from state-of-the-art for CVRP and VRPTW.}

    \item VRPSolver \citep{pessoa2020generic} is an exact, \added{state-of-the-art} VRP solver which supports different problem variants through a generic model.
    \added{The solver has a C++, Julia and Python interface, the latter of which can be installed directly from the Python package index.}
    VRPSolver can find optimal solutions and prove optimality for VRP solutions of modest size, but it is does not scale to instances with more than a few hundred customers.
    \added{Moreover, the solver's license limits the use of its most powerful components to academic users.}

    \item \added{
        ``A VRP Solver'' is a rich VRP heuristic solver due to \cite{builuk_rosomaxa_2023}, written in Rust and made available under an Apache 2.0 license.
        The project supports many VRP variants, using a custom data format based on JavaScript Object Notation.
        The project is well-tested and a user manual is available, including examples, but appears to be lacking a detailed documentation of the available functional endpoints.
        Furthermore, it is unclear how well the solver performs in general, as results on standard benchmark instances are lacking.
    }
\end{itemize}

While each of these projects has their own merit,\deleted{ many lack extensive documentation, tests or performance, limiting their ease of use.}
PyVRP has a unique combination of scope, performance, flexibility and ease-of-use, making it a useful addition to this set of projects.

\section{Technical implementation}
\label{sec:technical_implementation}

PyVRP implements a variant of the HGS algorithm of \cite{vidal2013hybrid}.
At its core, our implementation consists of a genetic algorithm, a population, and a local search \replaced{improvement method}{educator}.
We explain these below, but refer to our documentation at \url{https://pyvrp.org/} for a full overview of the class and function descriptions, and other helper classes and methods we do not describe here.

\subsection{Overview of HGS}
HGS is a hybrid algorithm that combines a genetic algorithm with a local search algorithm.
It maintains a population with feasible and infeasible solutions.
Initially, solutions are created by randomly assigning customers to routes (feasibility is not required), which ensures diversity in the search.
Then, in every iteration, two parents are selected from the population, and combined using a crossover operator to create a new \emph{offspring solution}.
We provide an efficient C++ implementation of the selective route exchange (SREX) crossover operator~\citep{nagata2010memetic} by default, but this can easily be \replaced{replaced by}{switched out for} another crossover operator.

In each iteration, the new offspring solution is improved using local search, which considers time windows and capacities as soft constraints by penalising violations.
This way, the local search considers a smoothed version of the problem, which helps the genetic algorithm to converge towards promising regions of the solution space.
The penalty weights are automatically adjusted such that a target percentage of the local search runs results in a feasible solution. 
After the local search, the offspring is inserted into the population.
Once the population exceeds a certain size, a survivor selection mechanism removes solutions which contribute the least to the overall quality and diversity of the population.

\subsection{Genetic algorithm}
The genetic algorithm is implemented in Python and defines the main search loop.
In every iteration of the search loop, the genetic algorithm selects two (feasible or infeasible) parent solutions from the population.
A crossover operators takes the two parent solutions and uses those to generate an offspring solution that inherits features from both parents.

After crossover completes, the offspring solution is improved using local search, and then added to the population.
If this improved offspring solution is feasible and better than our current best solution, it becomes the new best observed solution.
Finally, after the main search loop completes, the genetic algorithm returns a result object that contains the best observed solution and detailed runtime statistics.

\subsection{Local search}
\label{subsec:local_search}

We provide an efficient local search implementation to improve a new offspring solution.
\added{This improvement procedure is typically the most expensive part of the HGS algorithm.
Software profiling suggests that in PyVRP it accounts for 80-90\% of the runtime, which is why the local search is implemented in C++.}
The implementation explores a granular neighbourhood \citep{toth2003granular} in a very efficient manner using user-provided operators.
These operators evaluate moves in different neighbourhoods, and the local search algorithm applies the move as soon as it yields a direct improvement in the objective value of the solution.
The search is repeated until no more improvements can be made. 
We distinguish \emph{node operators} and \emph{route operators}.
Node operators are applied to pairs of customers, and evaluate local moves around these customers.
\added{Node operators may also be applied to a customer and an unassigned vehicle: this evaluates moves placing a customer into an empty route.
To limit the number of vehicles used, these moves are \replaced{evaluated}{checked} only once all moves involving pairs of customer have been exhausted.}
Route operators are applied to pairs of \added{non-empty} routes and evaluate more expensive moves that intensify the search.

Users are free to supply their own node and route operators, but for convenience we already provide a large set of \deleted{very }efficient operators\replaced{, which we describe next}{.
We describe these next}.

\subsubsection{Node operators}
Node operators each evaluate (and possibly apply) a move between two customers $u$ and $v$, with the restriction that $v$ is in the granular neighbourhood $\mathcal{N}(u)$ of $u$.
We provide a default granular neighbourhood of size $k$ for each customer, that takes into account both spatial and temporal aspects of the problem instance.
This default implementation reduces the neighborhood size from $O(n^2)$ to $O(kn)$, but a user can fully customise the neighbourhood structure, or replace it altogether with their own.

\replaced{PyVRP currently implements the following node operators}{The node operators that are implemented in PyVRP are}:

\paragraph{$(N, M)$-exchange} 
This operator considers exchanging a consecutive route segment of $N > 0$ nodes starting at $u$ (inclusive) with a segment of $0 \le M \le N$ nodes starting at $v$ (inclusive).
These segments must not overlap in the same route\added{, and not contain the depot}.
When $M=0$, this operator evaluates \emph{relocate} moves inserting $u$ (and possible subsequent nodes) after $v$.
When $M > 0$, the operator evaluates \emph{swap} moves exchanging the route segments of one or more nodes starting at nodes $u$ and $v$.
This exchange generalises the implementations of \citet{vidal2022hybrid}.
We implement $(N, M)$-exchange using C++'s template mechanism, which after compilation results in efficient, specialised operator implementations for any $N$ and $M$.

\paragraph{MoveTwoClientsReversed}
This operator considers a $(2,0)$-exchange where $u$ and its immediate successor are reversed \added{before inserting them after $v$}. 

\paragraph{2-OPT}
The 2-OPT operator represents the routes of $u$ and $v$ as (directed) line-graphs, where an arc $u \rightarrow x$ indicates $x$ is visited \added{directly} after $u$. 
2-OPT replaces the arcs $u \rightarrow x$ and $v \rightarrow y$ by $u \rightarrow y$ and $v \rightarrow x$, effectively recombining the starts and ends of the two routes if we split them at $u$ and $v$.
When $u$ and $v$ are within the same route, and $u$ precedes $v$, this operator replaces $u \rightarrow x$ and $v \rightarrow y$ by $u \rightarrow v$ and $x \rightarrow y$, thus reversing the route segment from $x$ to $v$.

\subsubsection{Route operators}
Route operators consider moves between route pairs, avoiding the granularity restrictions imposed on the node operators.
This enables the evaluation of much larger neighbourhoods, while additional caching opportunities ensure these evaluations remain fast.
PyVRP provides two route operators by default:

\paragraph{RELOCATE*}
The RELOCATE* operator finds and applies the best $(1, 0)$-exchange move between two routes.
\added{RELOCATE* uses the $(N, M)$-exchange node operator (with $N = 1$ and $M = 0$) to evaluate each move between the two routes.}

\paragraph{SWAP*}
The SWAP* operator due to \cite{vidal2022hybrid} considers the best swap move between two routes, but does not require that the swapped customers are inserted in each others place.
Instead, each is inserted into the best location in the other route.
We enhance the implementation of \cite{vidal2022hybrid} with time window support, further caching, and earlier stopping when evaluating `known-bad' moves. 

\subsection{Population management}

The population is implemented in Python, using feasible and infeasible sub-populations that are implemented in C++ for performance.
New solutions can be added to the population, and parent solutions can be requested from it for crossover.
These parents are selected by a \replaced{$k$-way}{binary} tournament on the relative fitness of each parent~\added{\citep{Team_SB}}.
\added{By default, $k=2$, which results in a binary tournament.}

The population is initialised with a minimal set of random solutions.
New solutions obtained by the genetic algorithm are added to it as they are generated.
Once a sub-population reaches its maximal size, survivor selection is performed that reduces the sub-population to its minimal size.
This survivor selection is done by first removing duplicate solutions, and then by removing those solutions that have worst fitness based on the biased fitness criterion of \cite{vidal2022hybrid}.
This fitness criterion balances solution quality based on the solution's objective value and diversity w.r.t. to other solutions in the sub-population, evaluated using a diversity measure supplied to the population.
We implement a directed variant of the broken pairs distance, but a user can also supply their own diversity measure.

\section{The PyVRP package}
\label{sec:package}

The PyVRP package is developed in a GitHub repository located at \url{https://github.com/PyVRP/PyVRP}. 
The repository contains the C++ and Python source code, including unit and integration tests, as well as documentation and examples introducing new users to PyVRP.
Additionally, the repository uses automated workflows that build PyVRP for different platforms (currently Linux, Windows, and Mac OS), such that a user can install PyVRP directly from the Python package index using \texttt{pip install pyvrp} without having to compile the C++ extensions themselves.

\subsection{Package structure}
The top-level \texttt{pyvrp} namespace contains some of the components of Section~\ref{sec:technical_implementation}, and important additional classes.
These include \added{the \texttt{Model} modelling interface,} the \texttt{GeneticAlgorithm} and \texttt{Population}, along with a \texttt{read} function that can be used to read benchmark instances in various formats (through the VRPLIB Python package).
Crossover operators that can be used together with the \texttt{GeneticAlgorithm} are provided in \texttt{pyvrp.crossover}.
Further, the \texttt{pyvrp.diversity} namespace constains diversity measures that can be used with the \texttt{Population}.
The \texttt{pyvrp.\replaced{search}{educate}} namespace contains the \texttt{LocalSearch} class, the \deleted{node and route} operators, and the \texttt{compute\_neighbours} function that computes a granular neighbourhood.
Stopping criteria for the genetic algorithm are provided by \texttt{pyvrp.stop}.
These include stopping criteria based on a maximum number of iterations or runtime, but also variants that stop after a number of iterations without improvement.
Finally, \texttt{pyvrp.plotting} provides utilities for plotting and analysing solutions.

\subsection{Example use}
\added{
We present two examples, for different audiences.
The first example in Listing~\ref{lst:model_api} shows the modelling interface of PyVRP, and how that can be used to define and solve a CVRP instance.
This interface is particularly convenient for practitioners interested in solving VRPs using PyVRP.
The second example in Listing~\ref{lst:pyvrp} shows the different components in PyVRP, and how they can be used to solve a VRPTW instance.
This example is helpful for understanding how PyVRP's implementation of HGS works, and can be used as a basis to customise the solution algorithm.
}

\added{
We will first present the modelling interface in Listing~\ref{lst:model_api}.
}

\begin{lstlisting}[style=mypython, language=Python, caption=Using PyVRP's modelling interface to solve a CVRP instance., label=lst:model_api]
import numpy as np
from pyvrp import Model
from pyvrp.stop import MaxRuntime


gen = np.random.default_rng(seed=42)  # random coordinate and demand data
coords = gen.integers(0, 100, size=(10, 2))
demands = gen.integers(0, 10, size=(10,))

m = Model()
m.add_vehicle_type(capacity=15, num_available=4)  # 4 vehicles of 15 capacity each
depot = m.add_depot(x=coords[0][0], y=coords[0][1])
clients = [
    m.add_client(x=coords[idx][0], y=coords[idx][1], demand=demands[idx])
    for idx in range(1, len(coords))
]

for frm in m.locations:
    for to in m.locations:
        distance = abs(frm.x - to.x) + abs(frm.y - to.y)  # Manhattan
        m.add_edge(frm, to, distance=distance)

res = m.solve(stop=MaxRuntime(1), seed=4)  # one second
print(res)
\end{lstlisting}

\added{
The modelling interface is available as \texttt{Model}, and can be used to define all relevant instance attributes: the depot, clients, vehicle types, and the edges connecting all locations.
After defining an instance, it can be solved by calling the \texttt{solve} method on the model.
Once solving finishes, a result object \texttt{res} is returned.
This object contains the best-found solution (\texttt{res.best}) and statistics about the solver run.
The result object can be printed to display the solution and some relevant statistics.
Additionally, the results can be plotted, which we will show how to do in Listing~\ref{lst:pyvrp}.
}

In Listing~\ref{lst:pyvrp} we solve the 1000 customer \texttt{RC2\_10\_5} instance of the Homberger and Gehring VRPTW set of benchmarks.
\added{Rather than using the modelling interface's high-level \texttt{solve} method, here we set everything up explicitly.}
The code assumes that the \texttt{RC2\_10\_5} instance is available locally.

\begin{lstlisting}[style=mypython, language=Python, caption=PyVRP example usage., label=lst:pyvrp]
from pyvrp import *
from pyvrp.crossover import selective_route_exchange as srex
from pyvrp.diversity import broken_pairs_distance as bpd
from pyvrp.search import *
from pyvrp.plotting import plot_result
from pyvrp.stop import MaxRuntime

data = read("RC2_10_5.txt", instance_format="solomon", round_func="dimacs")
rng = RandomNumberGenerator(seed=42)

# We use local search with all default node and route operators to improve
# offspring after crossover.
ls = LocalSearch(data, rng, compute_neighbours(data))
for op in NODE_OPERATORS:
    ls.add_node_operator(op(data))
for op in ROUTE_OPERATORS:
    ls.add_route_operator(op(data))

pen_manager = PenaltyManager()
pop = Population(bpd)
init_pop = [Solution.make_random(data, rng) for _ in range(25)]
algo = GeneticAlgorithm(data, pen_manager, rng, pop, ls, srex, init_pop)

res = algo.run(stop=MaxRuntime(60))  # run for sixty seconds
plot_result(res, data)
\end{lstlisting}

\begin{figure}
    \centering
    \includegraphics[width=\textwidth]{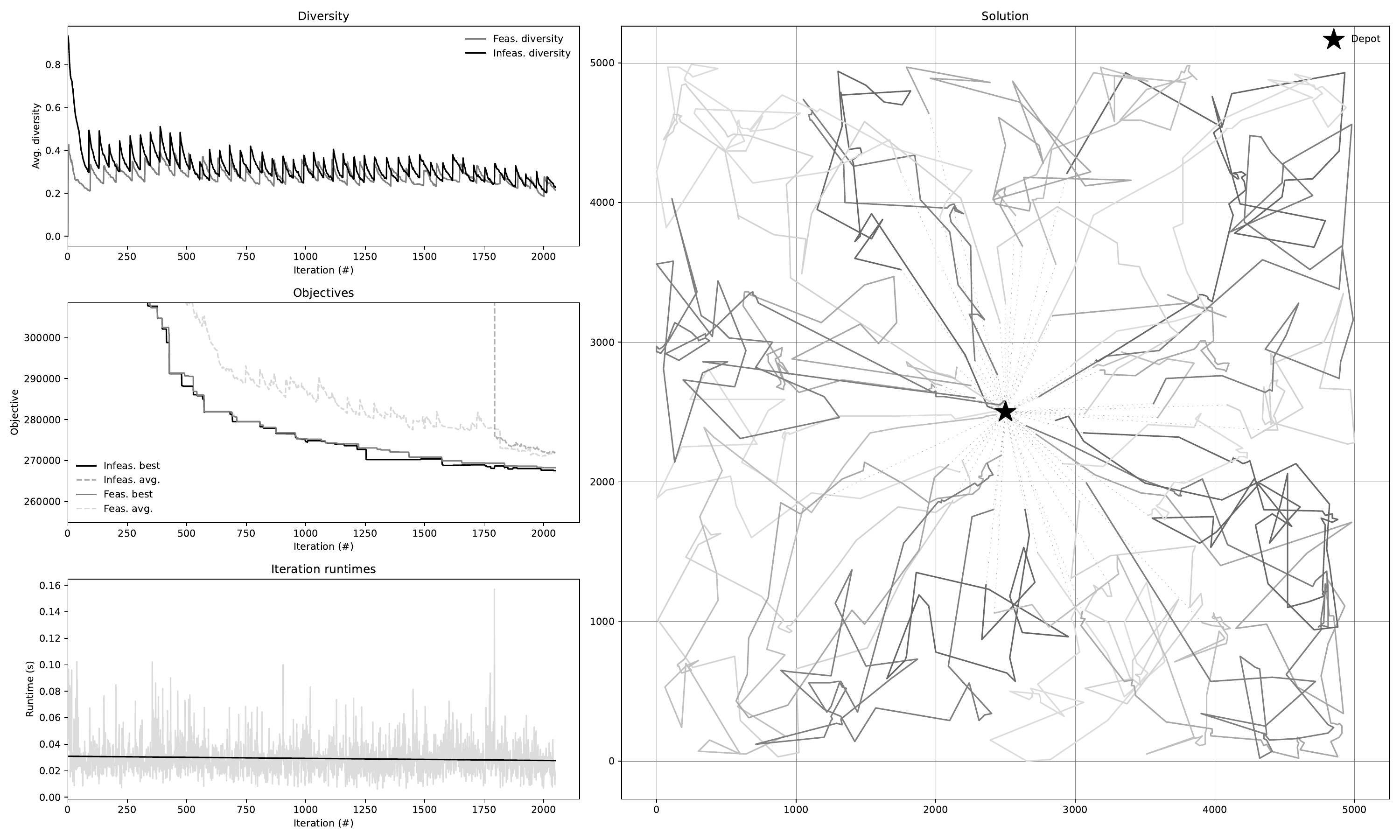}
    \caption{
        Detailed statistics collected from a single run of Listing \ref{lst:pyvrp}. 
        The top-left figure shows the average diversity of the feasible and infeasible sub-populations.
        It is clear from this figure that periodic survivor selection improves diversity.
        The middle-left figure shows the best and average objectives of both sub-populations, which improve over time as the search progresses.
        The bottom-left figure shows iteration runtimes (in seconds), \added{including a trendline}.
        Finally, the figure on the right plots the best observed solution.
    }
    \label{fig:RC2_10_5}
\end{figure}

\added{
Listing~\ref{lst:pyvrp} first reads a benchmark instance in standard format and constructs a random number generator with fixed seed.
It then defines the local search method.
We use the default granular neighbourhood computed by \texttt{compute\_neighbours}, but this can easily be customised by providing an alternative neighbourhood definition.
Then, we add all node and route operators described in Section~\ref{subsec:local_search} to the local search object.
This is not required: any subset of these operators is also allowed, and might even improve the solver performance in specific cases.
Finally, the penalty manager and population are initialised.
These track, respectively, the weights of constraint violation penalties, and the feasible and infeasible solution subpopulations.
An initial population should also be provided to the genetic algorithm: here we generate 25 random solutions.
A user may wish to apply alternative population generation methods here.
Finally, the genetic algorithm is initialised and run until a stopping criterion is met: in this case, the stopping criterion is 60 seconds of runtime.
We plot the solver trajectory and best observed solution in Figure~\ref{fig:RC2_10_5}.
}

\subsection{\added{Extending PyVRP}}

\added{
Before writing new code for PyVRP, a few things must be decided about the new constraint.
Hard constraints might require changes to PyVRP data structures.
Soft constraints typically require modifications to the cost evaluation functions.
Additionally, the new constraint likely requires additional data attributes that must be added to PyVRP's data instance object, and solution representation.
Once that new data is available, the search method can be updated to compute the correct cost deltas of each available move.
Some of the cost delta evaluation may need to be cached to ensure an efficient implementation---this is particularly the case for time-related costs, which PyVRP already supports.
Entirely new problem aspects might need to develop such caching as part of the extension.
}

\added{
Since we have developed several extensions to PyVRP already, there are some examples available of previous work.
We have summarised guidelines for extending PyVRP in our online documentation, available at \protect\url{https://pyvrp.org/dev/new_vrp_variants.html}.
}

\section{Experiments}
\label{sec:experiments}

In this section we present PyVRP's performance on widely used CVRP and VRPTW benchmark instances.
\added{We compare PyVRP's performance with both the best known solutions (BKSs), and with results from the literature, accounting for CPU differences by adjusting the time limits based on the PassMark score.}
\replaced{PyVRP is benchmarked}{All our benchmarks are performed} on an \replaced{AMD EPYC 7H12}{Intel Xeon Gold 5118} CPU with a PassMark single thread performance of \replaced{2014}{1807}.
\added{We benchmark PyVRP version 0.5.0, which is available as a static archive on the IJOC GitHub repository~\citep{wouda_et_al_2023_code_repo}.}
The BKSs were obtained from the CVRPLIB repository on 28 February 2023.

\subsection{CVRP}
We evaluate our solver for CVRP on the X benchmark instances of \cite{uchoa2017new}.
This set includes 100 instances and covers diverse problem characteristics, such as customer geography, demand distributions and route lengths.
We follow the convention to minimise the total distance.
The distances are computed by taking the Euclidean distances rounded to the nearest integer.
The parameter settings used in our experiments are shown in Table~\ref{tab:parameters} in Appendix~\ref{app:parameters}.
We solve each instance with ten different seeds and present the average total distance rounded to one decimal.
We compare our solver \replaced{with}{against} the results of state-of-the-art CVRP solvers HGS-2012 \citep{vidal2013hybrid} and HGS-CVRP \citep{vidal2022hybrid}.
We use the time limits of \cite{vidal2022hybrid}: each instance is solved for $T_{\max} = n \times 240/100$ seconds, where $n$ denotes the number of customers in the instance.
The smallest instance of 100 customers is thus solved for 4 minutes, and the largest instance with 1000 customers is solved for 40 minutes. 
The experiments from \cite{vidal2022hybrid} were run on a single core of an Intel Gold 6148 CPU with PassMark single thread performance of 2183. 
We normalise the time limit by multiplying $T_{\max}$ with 2183 / \replaced{2014}{1807} to compensate for the CPU differences.

\begin{table}
    \centering
    \caption{Benchmark results for CVRP on X instances of \citet{uchoa2017new}.}
    \begin{tabular}{lccc}
\toprule
 & Mean cost & Mean gap & Gap of mean \\ \midrule
PyVRP & \replaced{63275.5}{63294.0} & 0.22 \% & \replaced{0.27}{0.30} \% \\
HGS-2012 & 63285.8 & 0.21 \% & 0.28 \% \\
HGS-CVRP & 63206.1 & 0.11 \% & 0.16 \% \\
BKS & 63106.7 & 0.00 \% & 0.00 \%
     \\ \bottomrule
\end{tabular}

    \label{tab:cvrp-results-summary}
\end{table}

Table~\ref{tab:cvrp-results-summary} presents the summarised results for CVRP.
We report both the \emph{mean gap} $\frac{1}{n}\sum_i \frac{c_i}{\text{BKS}_i} - 1$, as well as the \emph{gap of the mean} $\frac{\frac{1}{n}\sum_i c_i}{\frac{1}{n}\sum_i \text{BKS}_i} - 1$. 
Here $i$ represents a single instance, $c_i$ denotes the objective value of the solver's solution, and $\text{BKS}_i$ the best known solution value. 
The gaps are expressed as percentages.
PyVRP obtains a mean gap of 0.22\% and a gap of the mean of \replaced{0.27}{0.30}\% on the solved instances. 
Despite the fact that PyVRP has not been specifically designed for the CVRP, these gaps are only slightly higher than the gaps of specialised CVRP solvers like HGS-2012 (mean gap 0.21\% and gap of mean 0.28 \%) and HGS-CVRP (mean gap 0.11\% and gap of mean 0.16\%). 
The complete results for each instance can be found in Table~\ref{tab:cvrp-results} in Appendix~\ref{app:cvrp-results}.

\subsection{VRPTW}
We evaluate our solver on the well-known Homberger and Gehring VRPTW instance set \citep{homberger1999two}.
The set includes five categories of instance sizes with 200, 400, 600, 800 and 1000 customers.
For brevity, we only presents results here for the 1000 customer instances.
The relative performance differences on smaller instances are similar to the results on the 1000 customer instances.
Following the DIMACS competition convention, we minimise the total travel distance and set a time limit of two hours for 1000 customer instances on a CPU with reference PassMark score of 2000.
We normalise the time limits for our choice of CPU, so we multiply the time limits by 2000 / \replaced{2014}{1807} in our numerical experiments.
We solve each instance with ten different seeds and present the average total distance rounded to 1 decimal.
The parameter settings of our solver are shown in Table~\ref{tab:parameters} and largely follow the parameter values used by \cite{kool2022hybrid} and \cite{van_doorn_solving_2022}.
We compare PyVRP \replaced{with}{against} the solutions found by \cite{kool2022hybrid}, denoted as HGS-DIMACS.
Since the original DIMACS results were reported as gaps to a reference solution provided by the DIMACS organisation, we also include the gaps \replaced{between these reference solutions and the reference BKSs}{with respect to these reference solutions}.

\begin{table}
    \centering
    \caption{Benchmark results for VRPTW on 1000 customer instances of \citet{homberger1999two}.}
    \begin{tabular}{lccc}
\toprule
 & Mean cost & Mean gap & Gap of mean \\ \midrule
PyVRP & \replaced{33296.4}{33321.6} & \replaced{0.40}{0.45}\% & \replaced{0.46}{0.54}\% \\
HGS-DIMACS & 33265.5 & 0.32\% & 0.37\% \\
DIMACS reference & 33245.1 & 0.29\% & 0.31\% \\
BKS & 33143.8 & 0.00\% & 0.00\%
     \\ \bottomrule
\end{tabular}

    \label{tab:vrptw-results-summary}
\end{table}

Table~\ref{tab:vrptw-results-summary} summarises the results for VRPTW. 
The complete results for each instance can be found in Table~\ref{tab:vrptw-results} in Appendix~\ref{app:vrptw-results}.
We again report both the mean gap and the gap of the mean.
PyVRP achieves a mean gap of \replaced{0.40}{0.45}\% and gap of mean of \replaced{0.46}{0.54}\% on the VRPTW benchmark instances, which is slightly higher than that of HGS-DIMACS (mean gap of 0.32\% and gap of mean 0.37\%).
The difference in performance can be explained by the simplified implementation of PyVRP.
As a result of this simplification, PyVRP would have ended up in second place in the DIMACS VRPTW competition.
\added{Furthermore, during extended runs, PyVRP managed to improve 27 of the 300 best known solutions of the complete Homberger and Gehring instances.}


\section{Conclusion}
\label{sec:conclusion}

We introduce PyVRP, an open-source Python package for solving the vehicle routing problem with time windows, and show numerically that PyVRP achieves excellent performance on this problem variant.
The package has minimal dependencies and can easily be installed in pre-compiled format from the Python package index.
The package's implementation is flexible, and can easily be extended with new solution techniques and vehicle routing problem variants.
Our hope is that PyVRP facilitates furthering the state-of-the-art in VRP solving by enabling researchers \added{and practitioners} to easily and quickly build on a state-of-the-art solver.

\ACKNOWLEDGMENT{%
We are grateful to Penghui Guo for transferring the `PyVRP' name on the Python package index to us. 
Leon Lan would like to thank TKI Dinalog, Topsector Logistics and the Dutch Ministry of Economic Affairs and Climate Policy for funding this project.
}

%
%
%

\begin{APPENDICES}
\section{Parameter settings}
\label{app:parameters}
The parameter settings used in our experiments are shown in Table~\ref{tab:parameters}.

\begin{table}
\centering
\caption{Parameters values for CVRP and VRPTW. The values are slightly adapted from those of \cite{vidal2022hybrid} for CVRP and \cite{kool2022hybrid} for VRPTW.}
\label{tab:parameters}
\begin{tabular}{@{}lp{.57\linewidth}cc@{}}
\toprule
Category                           & Parameter                                         & CVRP & VRPTW \\ \midrule
\multirow{3}{*}{Genetic algorithm} & Repair probability                                & 50\%  & 80\%  \\
                                   & Number of non-improving iterations before restart & 20000 & 20000 \\ 
\midrule
\multirow{6}{*}{Population}        & Minimum population size                           & 25    & 25    \\
                                   & Population generation size                        & 40    & 40    \\
                                   & Number of elite solutions                         & 4     & 4     \\
                                   & Number of close solutions                         & 5     & 5     \\
                                   & Lower bound diversity                             & 0.1   & 0.1   \\
                                   & Upper bound diversity                             & 0.5   & 0.5   \\ 
\midrule
\multirow{6}{*}{Penalty manager}   & Initial capacity penalty                          & 20    & 20    \\
                                   & Initial time warp penalty                         & -     & 6     \\
                                   & Repair booster                                    & 12    & 12    \\
                                   & Number of registrations between penalty updates   & 100    & 50    \\
                                   & Penalty increase factor                           & 1.25  & 1.34  \\
                                   & Penalty decrease                                  & 0.85  & 0.32  \\ 
                                   & Target feasible                                   & 0.43  & 0.43  \\
\midrule
\multirow{10}{*}{Local search}     & Number of neighbours                              & 20    & 40    \\
                                   & Weight waiting time                               & -     & 0.2   \\
                                   & Weight time warp                                  & -     & 1.0   \\
                                   & Symmetric proximity                               & True  & True  \\
                                   & Symmetric neighbours                              & True  & False \\
                                   & Relocate operators ($(1, 0)$-exchange, $(2, 0)$-exchange, \added{$(3, 0)$-exchange,} MoveTwoClientsReversed)                               & True  & True  \\
                                   & Swap operators ($(1, 1)$-exchange, $(2, 1)$-exchange, $(2, 2)$-exchange, $(3, 2)$-exchange, $(3, 3)$-exchange)                                   & True  & True  \\
                                   & Include 2-OPT operator                            & True  & True  \\
                                   & Include RELOCATE* operator                        & True  & True  \\
                                   & Include SWAP* operator                            & True  & True  \\
\bottomrule
\end{tabular}

\end{table}

\section{CVRP results}
\label{app:cvrp-results}
The complete CVRP benchmark results are presented in Table~\ref{tab:cvrp-results}.

\section{VRPTW results}
\label{app:vrptw-results}
The complete VRPTW benchmark results are presented in Table~\ref{tab:vrptw-results}.

\clearpage 

\renewcommand*{\arraystretch}{1.4}
\begin{longtable}{lcclcclcclc}
\caption{Benchmark results for CVRP on X instances.} \label{tab:cvrp-results}\\
\toprule
& \multicolumn{2}{c}{PyVRP} & \multicolumn{1}{l}{} & \multicolumn{2}{c}{HGS-2012} & \multicolumn{1}{l}{} & \multicolumn{2}{c}{HGS-CVRP} & \multicolumn{1}{l}{} & \multicolumn{1}{c}{BKS}  \\
\cmidrule{2-3}  \cmidrule{5-6} \cmidrule{8-9} \cmidrule{11-11}
Instance    & \multicolumn{1}{c}{Cost}  & \multicolumn{1}{c}{Gap} &  & \multicolumn{1}{c}{Cost}   & \multicolumn{1}{c}{Gap} &  & \multicolumn{1}{c}{Cost}             & \multicolumn{1}{c}{Gap} &  & \multicolumn{1}{c}{Cost} \\
\midrule 
\endfirsthead 
\caption{Benchmark results for CVRP on X instances (continued).} \\
\toprule
& \multicolumn{2}{c}{PyVRP} & \multicolumn{1}{l}{} & \multicolumn{2}{c}{HGS-2012} & \multicolumn{1}{l}{} & \multicolumn{2}{c}{HGS-CVRP} & \multicolumn{1}{l}{} & \multicolumn{1}{c}{BKS}  \\
\cmidrule{2-3}  \cmidrule{5-6} \cmidrule{8-9} \cmidrule{11-11}
Instance    & \multicolumn{1}{c}{Cost}  & \multicolumn{1}{c}{Gap} &  & \multicolumn{1}{c}{Cost}   & \multicolumn{1}{c}{Gap} &  & \multicolumn{1}{c}{Cost}             & \multicolumn{1}{c}{Gap} &  & \multicolumn{1}{c}{Cost} \\
\midrule
\endhead 
\bottomrule
\endfoot 
X-n101-k25  & 27591.0  & 0.00 &  & 27591.0  & 0.00 &  & 27591.0  & 0.00 &  & 27591  \\
X-n106-k14  & 26376.0  & 0.05 &  & 26408.8  & 0.18 &  & 26381.4  & 0.07 &  & 26362  \\
X-n110-k13  & 14971.0  & 0.00 &  & 14971.0  & 0.00 &  & 14971.0  & 0.00 &  & 14971  \\
X-n115-k10  & 12747.0  & 0.00 &  & 12747.0  & 0.00 &  & 12747.0  & 0.00 &  & 12747  \\
X-n120-k6   & 13332.0  & 0.00 &  & 13332.0  & 0.00 &  & 13332.0  & 0.00 &  & 13332  \\
X-n125-k30  & 55541.9  & 0.01 &  & 55539.0  & 0.00 &  & 55539.0  & 0.00 &  & 55539  \\
X-n129-k18  & 28940.0  & 0.00 &  & 28940.0  & 0.00 &  & 28940.0  & 0.00 &  & 28940  \\
X-n134-k13  & 10916.2  & 0.00 &  & 10916.0  & 0.00 &  & 10916.0  & 0.00 &  & 10916  \\
X-n139-k10  & 13590.0  & 0.00 &  & 13590.0  & 0.00 &  & 13590.0  & 0.00 &  & 13590  \\
X-n143-k7   & 15719.2  & 0.12 &  & 15700.0  & 0.00 &  & 15700.0  & 0.00 &  & 15700  \\
X-n148-k46  & 43448.0  & 0.00 &  & 43448.0  & 0.00 &  & 43448.0  & 0.00 &  & 43448  \\
X-n153-k22  & 21220.0  & 0.00 &  & 21223.5  & 0.02 &  & 21225.0  & 0.02 &  & 21220  \\
X-n157-k13  & 16876.0  & 0.00 &  & 16876.0  & 0.00 &  & 16876.0  & 0.00 &  & 16876  \\
X-n162-k11  & 14138.0  & 0.00 &  & 14138.0  & 0.00 &  & 14138.0  & 0.00 &  & 14138  \\
X-n167-k10  & 20557.0  & 0.00 &  & 20557.0  & 0.00 &  & 20557.0  & 0.00 &  & 20557  \\
X-n172-k51  & 45607.0  & 0.00 &  & 45607.0  & 0.00 &  & 45607.0  & 0.00 &  & 45607  \\
X-n176-k26  & 47817.0  & 0.01 &  & 47812.0  & 0.00 &  & 47812.0  & 0.00 &  & 47812  \\
X-n181-k23  & 25569.4  & 0.00 &  & 25570.2  & 0.00 &  & 25569.0  & 0.00 &  & 25569  \\
X-n186-k15  & 24149.6  & 0.02 &  & 24145.2  & 0.00 &  & 24145.0  & 0.00 &  & 24145  \\
X-n190-k8   & 16996.2  & 0.10 &  & 16992.4  & 0.07 &  & 16983.3  & 0.02 &  & 16980  \\
X-n195-k51  & 44236.1  & 0.03 &  & 44225.0  & 0.00 &  & 44225.0  & 0.00 &  & 44225  \\
X-n200-k36  & 58583.2  & 0.01 &  & 58589.6  & 0.02 &  & 58578.0  & 0.00 &  & 58578  \\
X-n204-k19  & 19568.5  & 0.02 &  & 19565.0  & 0.00 &  & 19565.0  & 0.00 &  & 19565  \\
X-n209-k16  & 30672.6  & 0.05 &  & 30658.7  & 0.01 &  & 30656.0  & 0.00 &  & 30656  \\
X-n214-k11  & 10874.6  & 0.17 &  & 10877.0  & 0.19 &  & 10860.5  & 0.04 &  & 10856  \\
X-n219-k73  & 117602.3 & 0.01 &  & 117601.7 & 0.01 &  & 117596.1 & 0.00 &  & 117595 \\
X-n223-k34  & 40486.3  & 0.12 &  & 40455.3  & 0.05 &  & 40437.0  & 0.00 &  & 40437  \\
X-n228-k23  & 25752.4  & 0.04 &  & 25742.7  & 0.00 &  & 25742.8  & 0.00 &  & 25742  \\
X-n233-k16  & 19238.4  & 0.04 &  & 19233.1  & 0.02 &  & 19230.0  & 0.00 &  & 19230  \\
X-n237-k14  & 27049.9  & 0.03 &  & 27049.4  & 0.03 &  & 27042.0  & 0.00 &  & 27042  \\
X-n242-k48  & 82875.4  & 0.15 &  & 82826.5  & 0.09 &  & 82806.0  & 0.07 &  & 82751  \\
X-n247-k50  & 37285.0  & 0.03 &  & 37295.0  & 0.06 &  & 37277.1  & 0.01 &  & 37274  \\
X-n251-k28  & 38769.4  & 0.22 &  & 38735.9  & 0.13 &  & 38689.9  & 0.02 &  & 38684  \\
X-n256-k16  & 18880.0  & 0.22 &  & 18880.0  & 0.22 &  & 18839.6  & 0.00 &  & 18839  \\
X-n261-k13  & 26604.5  & 0.18 &  & 26594.0  & 0.14 &  & 26558.2  & 0.00 &  & 26558  \\
X-n266-k58  & 75620.7  & 0.19 &  & 75646.8  & 0.22 &  & 75564.7  & 0.11 &  & 75478  \\
X-n270-k35  & 35315.2  & 0.07 &  & 35306.4  & 0.04 &  & 35303.0  & 0.03 &  & 35291  \\
X-n275-k28  & 21245.8  & 0.00 &  & 21247.8  & 0.01 &  & 21245.0  & 0.00 &  & 21245  \\
X-n280-k17  & 33592.6  & 0.27 &  & 33573.0  & 0.21 &  & 33543.2  & 0.12 &  & 33503  \\
X-n284-k15  & 20304.1  & 0.44 &  & 20248.0  & 0.16 &  & 20245.5  & 0.15 &  & 20215  \\
X-n289-k60  & 95304.3  & 0.16 &  & 95350.4  & 0.21 &  & 95300.9  & 0.16 &  & 95151  \\
X-n294-k50  & 47214.0  & 0.11 &  & 47217.8  & 0.12 &  & 47184.1  & 0.05 &  & 47161  \\
X-n298-k31  & 34250.5  & 0.06 &  & 34235.9  & 0.01 &  & 34234.8  & 0.01 &  & 34231  \\
X-n303-k21  & 21862.0  & 0.58 &  & 21763.4  & 0.13 &  & 21748.5  & 0.06 &  & 21736  \\
X-n308-k13  & 25895.4  & 0.14 &  & 25879.8  & 0.08 &  & 25870.8  & 0.05 &  & 25859  \\
X-n313-k71  & 94244.0  & 0.21 &  & 94127.7  & 0.09 &  & 94112.2  & 0.07 &  & 94043  \\
X-n317-k53  & 78360.8  & 0.01 &  & 78374.8  & 0.03 &  & 78355.4  & 0.00 &  & 78355  \\
X-n322-k28  & 29885.6  & 0.17 &  & 29887.5  & 0.18 &  & 29848.7  & 0.05 &  & 29834  \\
X-n327-k20  & 27620.3  & 0.32 &  & 27580.4  & 0.18 &  & 27540.8  & 0.03 &  & 27532  \\
X-n331-k15  & 31147.8  & 0.15 &  & 31114.0  & 0.04 &  & 31103.0  & 0.00 &  & 31102  \\
X-n336-k84  & 139627.1 & 0.37 &  & 139437.1 & 0.23 &  & 139273.5 & 0.12 &  & 139111 \\
X-n344-k43  & 42136.8  & 0.21 &  & 42086.0  & 0.09 &  & 42075.6  & 0.06 &  & 42050  \\
X-n351-k40  & 25998.8  & 0.40 &  & 25972.8  & 0.30 &  & 25943.6  & 0.18 &  & 25896  \\
X-n359-k29  & 51674.7  & 0.33 &  & 51653.8  & 0.29 &  & 51620.0  & 0.22 &  & 51505  \\
X-n367-k17  & 22827.2  & 0.06 &  & 22814.0  & 0.00 &  & 22814.0  & 0.00 &  & 22814  \\
X-n376-k94  & 147729.4 & 0.01 &  & 147719.0 & 0.00 &  & 147714.5 & 0.00 &  & 147713 \\
X-n384-k52  & 66155.7  & 0.35 &  & 66163.7  & 0.36 &  & 66049.1  & 0.18 &  & 65928  \\
X-n393-k38  & 38310.4  & 0.13 &  & 38281.4  & 0.06 &  & 38260.0  & 0.00 &  & 38260  \\
X-n401-k29  & 66264.0  & 0.17 &  & 66305.3  & 0.23 &  & 66252.5  & 0.15 &  & 66154  \\
X-n411-k19  & 19735.6  & 0.12 &  & 19723.8  & 0.06 &  & 19720.3  & 0.04 &  & 19712  \\
X-n420-k130 & 107903.6 & 0.10 &  & 107843.3 & 0.04 &  & 107839.8 & 0.04 &  & 107798 \\
X-n429-k61  & 65556.2  & 0.16 &  & 65565.4  & 0.18 &  & 65502.7  & 0.08 &  & 65449  \\
X-n439-k37  & 36422.6  & 0.09 &  & 36426.4  & 0.10 &  & 36395.5  & 0.01 &  & 36391  \\
X-n449-k29  & 55596.2  & 0.66 &  & 55598.1  & 0.66 &  & 55368.5  & 0.25 &  & 55233  \\
X-n459-k26  & 24191.1  & 0.22 &  & 24199.3  & 0.25 &  & 24163.8  & 0.10 &  & 24139  \\
X-n469-k138 & 222327.0 & 0.23 &  & 222364.3 & 0.24 &  & 222170.1 & 0.16 &  & 221824 \\
X-n480-k70  & 89600.2  & 0.17 &  & 89665.0  & 0.24 &  & 89524.4  & 0.08 &  & 89449  \\
X-n491-k59  & 66751.4  & 0.40 &  & 66723.7  & 0.36 &  & 66641.5  & 0.24 &  & 66483  \\
X-n502-k39  & 69252.4  & 0.04 &  & 69300.8  & 0.11 &  & 69239.5  & 0.02 &  & 69226  \\
X-n513-k21  & 24263.4  & 0.26 &  & 24206.5  & 0.02 &  & 24201.0  & 0.00 &  & 24201  \\
X-n524-k153 & 154881.0 & 0.19 &  & 154890.1 & 0.19 &  & 154747.6 & 0.10 &  & 154593 \\
X-n536-k96  & 95173.6  & 0.35 &  & 95205.1  & 0.38 &  & 95091.9  & 0.26 &  & 94846  \\
X-n548-k50  & 86855.6  & 0.18 &  & 86970.8  & 0.31 &  & 86778.4  & 0.09 &  & 86700  \\
X-n561-k42  & 42834.2  & 0.27 &  & 42783.9  & 0.16 &  & 42742.7  & 0.06 &  & 42717  \\
X-n573-k30  & 50925.0  & 0.50 &  & 50861.2  & 0.37 &  & 50813.0  & 0.28 &  & 50673  \\
X-n586-k159 & 190643.4 & 0.17 &  & 190759.3 & 0.23 &  & 190588.1 & 0.14 &  & 190316 \\
X-n599-k92  & 108813.2 & 0.33 &  & 108872.3 & 0.39 &  & 108656.0 & 0.19 &  & 108451 \\
X-n613-k62  & 59795.0  & 0.44 &  & 59801.0  & 0.45 &  & 59696.3  & 0.27 &  & 59535  \\
X-n627-k43  & 62439.1  & 0.44 &  & 62558.7  & 0.63 &  & 62371.6  & 0.33 &  & 62164  \\
X-n641-k35  & 63993.0  & 0.49 &  & 64086.0  & 0.63 &  & 63874.2  & 0.30 &  & 63682  \\
X-n655-k131 & 106851.9 & 0.07 &  & 106865.4 & 0.08 &  & 106808.8 & 0.03 &  & 106780 \\
X-n670-k130 & 146893.7 & 0.38 &  & 147319.0 & 0.67 &  & 146777.7 & 0.30 &  & 146332 \\
X-n685-k75  & 68532.3  & 0.48 &  & 68498.0  & 0.43 &  & 68343.1  & 0.20 &  & 68205  \\
X-n701-k44  & 82462.8  & 0.66 &  & 82457.9  & 0.65 &  & 82237.3  & 0.38 &  & 81923  \\
X-n716-k35  & 43616.9  & 0.56 &  & 43615.1  & 0.56 &  & 43505.8  & 0.31 &  & 43373  \\
X-n733-k159 & 136526.4 & 0.25 &  & 136512.5 & 0.24 &  & 136426.9 & 0.18 &  & 136187 \\
X-n749-k98  & 77801.3  & 0.69 &  & 77783.0  & 0.67 &  & 77655.4  & 0.50 &  & 77269  \\
X-n766-k71  & 115135.3 & 0.63 &  & 114894.6 & 0.42 &  & 114764.5 & 0.30 &  & 114417 \\
X-n783-k48  & 72915.8  & 0.73 &  & 73027.6  & 0.89 &  & 72790.7  & 0.56 &  & 72386  \\
X-n801-k40  & 73655.2  & 0.48 &  & 73803.3  & 0.68 &  & 73500.4  & 0.27 &  & 73305  \\
X-n819-k171 & 158745.0 & 0.39 &  & 158756.1 & 0.40 &  & 158511.6 & 0.25 &  & 158121 \\
X-n837-k142 & 194291.2 & 0.29 &  & 194636.5 & 0.46 &  & 194231.3 & 0.26 &  & 193737 \\
X-n856-k95  & 89129.8  & 0.19 &  & 89216.1  & 0.28 &  & 89037.5  & 0.08 &  & 88965  \\
X-n876-k59  & 99824.3  & 0.53 &  & 99889.4  & 0.59 &  & 99682.7  & 0.39 &  & 99299  \\
X-n895-k37  & 54281.0  & 0.78 &  & 54255.9  & 0.74 &  & 54070.6  & 0.39 &  & 53860  \\
X-n916-k207 & 329935.4 & 0.23 &  & 330234.0 & 0.32 &  & 329852.0 & 0.20 &  & 329179 \\
X-n936-k151 & 133365.1 & 0.49 &  & 133613.7 & 0.68 &  & 133369.9 & 0.49 &  & 132715 \\
X-n957-k87  & 85678.0  & 0.25 &  & 85823.3  & 0.42 &  & 85550.1  & 0.10 &  & 85465  \\
X-n979-k58  & 119781.8 & 0.68 &  & 119502.3 & 0.44 &  & 119247.5 & 0.23 &  & 118976 \\
X-n1001-k43 & 73001.0  & 0.89 &  & 73051.4  & 0.96 &  & 72748.8  & 0.54 &  & 72355  \\
\midrule
Mean & 63275.5 & 0.22 & & 63285.8 & 0.21 & & 63206.1 & 0.11 & & \multicolumn{1}{r}{63106.7} \\
Gap of mean & \multicolumn{1}{l}{} & 0.27 & & \multicolumn{1}{l}{} & 0.28 & & \multicolumn{1}{l}{} & 0.16 & &
\end{longtable}
\renewcommand*{\arraystretch}{1.4}
\begin{longtable}{lcclcclcclc}
\caption{Benchmark results for VRPTW on 1000-customer Homberger and Gehring instances.} 
\label{tab:vrptw-results}\\ 
\toprule
& \multicolumn{2}{c}{PyVRP} & \multicolumn{1}{l}{} & \multicolumn{2}{c}{HGS-DIMACS} & \multicolumn{1}{l}{} & \multicolumn{2}{c}{DIMACS} & \multicolumn{1}{l}{} & \multicolumn{1}{c}{BKS}  \\
\cmidrule{2-3}  \cmidrule{5-6} \cmidrule{8-9} \cmidrule{11-11}
Instance    & \multicolumn{1}{c}{Cost}  & \multicolumn{1}{c}{Gap} &  & \multicolumn{1}{c}{Cost}   & \multicolumn{1}{c}{Gap} &  & \multicolumn{1}{c}{Cost}             & \multicolumn{1}{c}{Gap} &  & \multicolumn{1}{c}{Cost} \\
\midrule
\endfirsthead 
\caption{Benchmark results for VRPTW on 1000-customer Homberger and Gehring instances (continued).}\\
\toprule
& \multicolumn{2}{c}{PyVRP} & \multicolumn{1}{l}{} & \multicolumn{2}{c}{HGS-DIMACS} & \multicolumn{1}{l}{} & \multicolumn{2}{c}{DIMACS} & \multicolumn{1}{l}{} & \multicolumn{1}{c}{BKS}  \\
\cmidrule{2-3}  \cmidrule{5-6} \cmidrule{8-9} \cmidrule{11-11}
Instance    & \multicolumn{1}{c}{Cost}  & \multicolumn{1}{c}{Gap} &  & \multicolumn{1}{c}{Cost}   & \multicolumn{1}{c}{Gap} &  & \multicolumn{1}{c}{Cost}             & \multicolumn{1}{c}{Gap} &  & \multicolumn{1}{c}{Cost} \\
\midrule
\endhead 
\bottomrule
\endfoot 
\bottomrule
\endlastfoot 
C1\_10\_1   & 42444.8 & 0.00 &  & 42444.8 & 0.00 &  & 42444.8 & 0.00 &  & 42444.8 \\
C1\_10\_2   & 41391.0 & 0.13 &  & 41386.4 & 0.12 &  & 41392.7 & 0.13 &  & 41337.8 \\
C1\_10\_3   & 40205.4 & 0.35 &  & 40216.4 & 0.38 &  & 40146.2 & 0.20 &  & 40064.4 \\
C1\_10\_4   & 39544.2 & 0.28 &  & 39524.6 & 0.23 &  & 39490.9 & 0.14 &  & 39434.1 \\
C1\_10\_5   & 42434.8 & 0.00 &  & 42434.8 & 0.00 &  & 42434.8 & 0.00 &  & 42434.8 \\
C1\_10\_6   & 42437.0 & 0.00 &  & 42437.0 & 0.00 &  & 42437.0 & 0.00 &  & 42437.0 \\
C1\_10\_7   & 42420.5 & 0.00 &  & 42420.4 & 0.00 &  & 42420.4 & 0.00 &  & 42420.4 \\
C1\_10\_8   & 41870.3 & 0.52 &  & 41902.5 & 0.60 &  & 41837.8 & 0.45 &  & 41652.1 \\
C1\_10\_9   & 40520.8 & 0.58 &  & 40425.2 & 0.34 &  & 40366.7 & 0.19 &  & 40288.4 \\
C1\_10\_10  & 40134.9 & 0.80 &  & 40149.5 & 0.84 &  & 39874.5 & 0.14 &  & 39816.8 \\
C2\_10\_1   & 16841.1 & 0.00 &  & 16841.1 & 0.00 &  & 16841.1 & 0.00 &  & 16841.1 \\
C2\_10\_2   & 16464.8 & 0.01 &  & 16462.6 & 0.00 &  & 16462.6 & 0.00 &  & 16462.6 \\
C2\_10\_3   & 16041.7 & 0.03 &  & 16036.5 & 0.00 &  & 16036.5 & 0.00 &  & 16036.5 \\
C2\_10\_4   & 15484.2 & 0.16 &  & 15463.6 & 0.03 &  & 15482.9 & 0.15 &  & 15459.5 \\
C2\_10\_5   & 16522.0 & 0.00 &  & 16521.3 & 0.00 &  & 16521.3 & 0.00 &  & 16521.3 \\
C2\_10\_6   & 16296.1 & 0.03 &  & 16290.7 & 0.00 &  & 16290.7 & 0.00 &  & 16290.7 \\
C2\_10\_7   & 16378.9 & 0.00 &  & 16378.4 & 0.00 &  & 16378.4 & 0.00 &  & 16378.4 \\
C2\_10\_8   & 16035.5 & 0.04 &  & 16029.8 & 0.00 &  & 16029.1 & 0.00 &  & 16029.1 \\
C2\_10\_9   & 16089.1 & 0.09 &  & 16075.6 & 0.00 &  & 16077.0 & 0.01 &  & 16075.4 \\
C2\_10\_10  & 15741.0 & 0.08 &  & 15728.6 & 0.00 &  & 15728.6 & 0.00 &  & 15728.6 \\
R1\_10\_1   & 53378.9 & 0.63 &  & 53341.7 & 0.56 &  & 53233.9 & 0.35 &  & 53046.5 \\
R1\_10\_2   & 48684.4 & 0.87 &  & 48513.0 & 0.52 &  & 48369.1 & 0.22 &  & 48263.1 \\
R1\_10\_3   & 45057.2 & 0.85 &  & 44999.1 & 0.72 &  & 44862.4 & 0.41 &  & 44677.1 \\
R1\_10\_4   & 42839.8 & 0.94 &  & 42625.9 & 0.44 &  & 42577.9 & 0.32 &  & 42440.7 \\
R1\_10\_5   & 50572.4 & 0.33 &  & 50596.5 & 0.38 &  & 50487.8 & 0.16 &  & 50406.7 \\
R1\_10\_6   & 47312.8 & 0.82 &  & 47276.2 & 0.74 &  & 47145.8 & 0.46 &  & 46930.3 \\
R1\_10\_7   & 44342.7 & 0.78 &  & 44174.8 & 0.40 &  & 44231.4 & 0.53 &  & 43997.4 \\
R1\_10\_8   & 42611.3 & 0.79 &  & 42451.6 & 0.41 &  & 42435.9 & 0.37 &  & 42279.3 \\
R1\_10\_9   & 49387.7 & 0.46 &  & 49417.9 & 0.52 &  & 49334.3 & 0.35 &  & 49162.8 \\
R1\_10\_10  & 47705.7 & 0.72 &  & 47690.5 & 0.69 &  & 47570.6 & 0.43 &  & 47364.6 \\
R2\_10\_1   & 36932.0 & 0.14 &  & 36891.3 & 0.03 &  & 36899.4 & 0.05 &  & 36881.0 \\
R2\_10\_2   & 31349.5 & 0.34 &  & 31384.7 & 0.46 &  & 31296.2 & 0.17 &  & 31241.9 \\
R2\_10\_3   & 24469.2 & 0.29 &  & 24429.7 & 0.13 &  & 24422.1 & 0.09 &  & 24399.0 \\
R2\_10\_4   & 17926.3 & 0.64 &  & 17870.1 & 0.33 &  & 17968.1 & 0.88 &  & 17811.5 \\
R2\_10\_5   & 34181.1 & 0.14 &  & 34195.3 & 0.18 &  & 34233.0 & 0.29 &  & 34132.8 \\
R2\_10\_6   & 29239.6 & 0.39 &  & 29179.4 & 0.19 &  & 29215.3 & 0.31 &  & 29124.7 \\
R2\_10\_7   & 23237.3 & 0.58 &  & 23305.2 & 0.88 &  & 23243.8 & 0.61 &  & 23102.2 \\
R2\_10\_8   & 17505.4 & 0.58 &  & 17463.3 & 0.34 &  & 17512.7 & 0.63 &  & 17403.8 \\
R2\_10\_9   & 32089.8 & 0.31 &  & 32058.0 & 0.21 &  & 32104.9 & 0.36 &  & 31990.6 \\
R2\_10\_10  & 29934.0 & 0.31 &  & 29882.3 & 0.14 &  & 29956.9 & 0.39 &  & 29840.5 \\
RC1\_10\_1  & 46018.1 & 0.50 &  & 46047.3 & 0.56 &  & 45948.5 & 0.34 &  & 45790.8 \\
RC1\_10\_2  & 44078.7 & 0.92 &  & 43892.6 & 0.49 &  & 43870.6 & 0.44 &  & 43678.3 \\
RC1\_10\_3  & 42537.9 & 0.99 &  & 42311.1 & 0.45 &  & 42338.0 & 0.51 &  & 42122.0 \\
RC1\_10\_4  & 41669.5 & 0.75 &  & 41577.1 & 0.53 &  & 41469.2 & 0.27 &  & 41357.4 \\
RC1\_10\_5  & 45333.7 & 0.68 &  & 45349.6 & 0.71 &  & 45235.5 & 0.46 &  & 45028.1 \\
RC1\_10\_6  & 45176.9 & 0.61 &  & 45359.9 & 1.02 &  & 45144.0 & 0.54 &  & 44903.6 \\
RC1\_10\_7  & 44742.2 & 0.73 &  & 44657.9 & 0.54 &  & 44686.2 & 0.61 &  & 44417.1 \\
RC1\_10\_8  & 44221.9 & 0.70 &  & 44145.6 & 0.52 &  & 44130.8 & 0.49 &  & 43916.5 \\
RC1\_10\_9  & 44153.4 & 0.67 &  & 44087.3 & 0.52 &  & 44072.4 & 0.49 &  & 43858.1 \\
RC1\_10\_10 & 43815.9 & 0.65 &  & 43635.1 & 0.23 &  & 43745.6 & 0.49 &  & 43533.7 \\
RC2\_10\_1  & 28168.0 & 0.16 &  & 28192.1 & 0.25 &  & 28190.8 & 0.24 &  & 28122.6 \\
RC2\_10\_2  & 24276.0 & 0.11 &  & 24268.6 & 0.08 &  & 24271.7 & 0.10 &  & 24248.6 \\
RC2\_10\_3  & 19673.1 & 0.28 &  & 19665.6 & 0.24 &  & 19747.9 & 0.66 &  & 19618.1 \\
RC2\_10\_4  & 15744.6 & 0.56 &  & 15716.2 & 0.38 &  & 15804.9 & 0.94 &  & 15657.0 \\
RC2\_10\_5  & 25851.7 & 0.21 &  & 25862.2 & 0.25 &  & 25892.3 & 0.37 &  & 25797.5 \\
RC2\_10\_6  & 25854.3 & 0.28 &  & 25823.0 & 0.16 &  & 25871.0 & 0.34 &  & 25782.5 \\
RC2\_10\_7  & 24458.4 & 0.26 &  & 24456.6 & 0.25 &  & 24451.7 & 0.23 &  & 24395.8 \\
RC2\_10\_8  & 23331.9 & 0.22 &  & 23307.9 & 0.12 &  & 23332.3 & 0.22 &  & 23280.2 \\
RC2\_10\_9  & 22784.8 & 0.23 &  & 22796.2 & 0.28 &  & 22859.3 & 0.56 &  & 22731.6 \\
RC2\_10\_10 & 21838.2 & 0.47 &  & 21862.3 & 0.58 &  & 21848.3 & 0.52 &  & 21736.1 \\
\midrule
Mean        & 33296.4 & 0.40 &  & 33265.5 & 0.32 &  & 33245.1 & 0.29 &  & 33143.8 \\
Gap of mean & \multicolumn{1}{l}{} & 0.46 & & \multicolumn{1}{l}{} & 0.37 & & \multicolumn{1}{l}{} & 0.31 & & \multicolumn{1}{l}{}
\end{longtable}

\end{APPENDICES}


\bibliographystyle{informs2014} 
\bibliography{references.bib} 


\end{document}